\documentclass{article}

\usepackage[preprint,nonatbib]{neurips_2019}

\usepackage[utf8]{inputenc} 
\usepackage[T1]{fontenc}    
\usepackage{hyperref}       
\usepackage{url}            
\usepackage{booktabs}       
\usepackage{amsfonts}       
\usepackage{nicefrac}       
\usepackage{microtype}      
\usepackage{graphicx}
\usepackage{subfig}

\title{Learnable Parameter Similarity}

\author{%
  Guangcong Wang \\
  Sun Yat-sen University\\
  \texttt{wanggc3@mail2.sysu.edu.cn} \\
  \And
  Jianhuang Lai\thanks{Corresponding author: Jianhuang Lai.} \\
  Sun Yat-sen University\\
  \texttt{stsljh@mail.sysu.edu.cn} \\
  \AND
  Wenqi Liang \\
  Sun Yat-sen University\\
  \texttt{liangwq8@mail2.sysu.edu.cn} \\
  \And
  Guangrun Wang \\
  Sun Yat-sen University\\
  \texttt{wanggrun@mail2.sysu.edu.cn} \\
}

\begin{document}

\maketitle

\begin{abstract}
  Most of the existing approaches focus on specific visual tasks while ignoring the relations between them. Estimating task relation sheds light on the learning of high-order semantic concepts, e.g., transfer learning. How to reveal the underlying relations between different visual tasks remains largely unexplored. In this paper, we propose a novel \textbf{L}earnable \textbf{P}arameter \textbf{S}imilarity (\textbf{LPS}) method that learns an effective metric to measure the similarity of second-order semantics hidden in trained models. LPS is achieved by using a second-order neural network to align high-dimensional model parameters and learning second-order similarity in an end-to-end way. In addition, we create a model set called ModelSet500 as a parameter similarity learning benchmark that contains 500 trained models. Extensive experiments on ModelSet500 validate the effectiveness of the proposed method. Code will be released at \url{https://github.com/Wanggcong/learnable-parameter-similarity}.
\end{abstract}

\section{Introduction}
Purpose-specific visual tasks have achieved greater commercial success by focusing on specific optimization problems, e.g., face recognition, object classification, object detection, visual object tracking, and instance segmentation. Can we exploit the underlying relations between different tasks and extend these task-specific methods to task-generic ones? Or can we connect shallow AI to general AI via task relations? Task relation sheds light on the learning of high-order semantic concepts.

Lots of evidence reveals transfer learning approaches \cite{pan2010survey,tzeng2017adversarial,liang2018m2m} that exploit the underlying relations between different tasks can further improve purpose-specific visual tasks with less labeled data. For instance, domain adaption methods \cite{tzeng2017adversarial,liang2018m2m} attempt to gain knowledge from source tasks and then apply it to a different but related target task. It is assumed that the knowledge learned from source tasks can help the learning of the target task.

Driven by these transfer learning methods, one would think: how to measure the relations between different tasks? Existing methods that offer partial solutions for this problem can be categorized into two groups. In the first group, a wide variety of transfer learning methods simply assume that tasks are related or unrelated based on human intuition or experience. For example, the knowledge gained from car recognition could be applied to truck recognition because cars intuitively look like trucks. However, one drawback of these methods is that human intuition could be different from machine learning principles. A negative transfer \cite{rosenstein2005transfer} could happen when human intuition is unreliable and the source domain data could lead to the reduced performance in the target domain. When the number of source domains is very large in some scenarios, it is hard to directly tell which is the best one for transfer learning.

In the second group, some methods attempt to jointly optimize multiple tasks and estimate task relations by cross-validation. For example, a taskonomy method \cite{zamir2018taskonomy} computes an affinity matrix among tasks based on whether the solution for one task can be sufficiently easily read out of the representation trained for another task. It uses $22\times 25$ transfer networks for the first-order transfer of 26 tasks. However, this pipeline requires a large amount of computation cost to jointly train all of the subsets of a task set. When a new task comes, it is needed to jointly train this new task and old tasks, which strongly limits its applications in real-world scenarios.

To address these drawbacks of existing methods, we propose a novel Learnable Parameter Similarity (LPS) method that learns second-order similarity to measure task relations by using trained task-specific models. Our observation is that the distance between intra-task models is closer than that between inter-task models. Let $\mathcal{T}$ denote a set of tasks. For each task $\mathcal{T}_i\in \mathcal{T}$, we repeat the training procedure $m_i$ times and thus obtain $M = \sum\limits {{m_i}}$ trained models. We then use these task-specific models as metadata points to train a second-order neural network to measure the parameter similarity.

Different from existing transfer learning methods, the LPS method measures task relation using task-specific models that are trained on independent task-specific datasets without jointly optimizing two many subsets of a task set. LPS pays attention to higher-order semantics/concepts, as illustrated in Figure \ref{fig:problem}. Data points produce a data similarity metric. Data similarity metrics produce a parameter similarity metric. If data is the zero-order similarity and data similarity is the first-order similarity, then parameter similarity can be regarded as the second-order similarity. LPS is also different from learning to learn methods. The former is to learn second-order similarity based on first-order similarity while the later is to learn hyper-parameters to know how to learn, which still focuses on the optimization of the first-order similarity.


Overall, the key contributions of this paper are:
\begin{itemize}
  \item We propose a novel Learnable Parameter Similarity (LPS) method that learns second-order similarity to measure task relation by using trained task-specific models instead of jointly training a large number of transfer networks.
  \item We introduce a hierarchical second-order network to deal with high-dimensional unaligned deep models and learn an effective parameter representation.
  \item We create a parameter similarity learning benchmark called ModelSet500 and extensive experiments on ModelSet500 validate the effectiveness of the proposed method.
\end{itemize}

\section{Related work}
\subsection{Transfer Learning}
Transfer Learning \cite{pan2010survey,bengio2011deep,long2017deep,rosenstein2005transfer,li2009transfer,tzeng2017adversarial,liang2018m2m,liu2016coupled,hoffman2017cycada} is to transfer knowledge from a source domain to a target domain, which has already achieved significant success in many areas including classification, regression, and clustering. Lots of approaches simply assume that source and target tasks are related or unrelated. For example, Tzeng et. al. \cite{tzeng2017adversarial} proposed to transfer the knowledge from RGB image based classification to depth image based classification. Liang et. al. \cite{liang2018m2m} proposed to transfer the person re-identification knowledge from one scene to another scene. In addition, some methods jointly optimize multiple tasks and estimate task relations by cross-validation. For example, Rosenstein et al. \cite{rosenstein2005transfer} proposed to detect and avoid negative transfer using very little data from the target task and empirically showed that dissimilar tasks may hurt the performance of the target task. Zamir et. al. \cite{zamir2018taskonomy} proposed to transfer multiple different tasks to target tasks based on an affinity matrix of transferabilities. These approaches either assume that those tasks that are intuitively related have a positive impact on the transfer learning or attempt to learn task relation by jointly optimizing multiple labeled datasets. Different from these methods, the proposed method learns task relation from trained task-specific models.

\subsection{Meta-learning/Learning to Learn}
Meta-learning, also known as learning to learn, is a subfield of machine learning that focuses on automated learning algorithms. Recently, lots of approaches \cite{zoph2016neural,andrychowicz2016learning,finn2017model,liu2018progressive,liu2018darts} aimed to automatically search hyper-parameters by using learning to learn models. For example, Zoph and Le \cite{zoph2016neural} used a recurrent network to generate the model descriptions of neural networks and trained the RNN with reinforcement learning to maximize the expected accuracy of the generated architectures on a validation set. Liu et. al. \cite{liu2018progressive} used a sequential model-based optimization strategy to search for structures in order of increasing complexity, while simultaneously learning a surrogate model to guide the search through structure space. Andrychowicz et. al. \cite{andrychowicz2016learning} attempted to learn an LSTM based neural optimizer to learn how to optimize neural networks. Finn et. al. \cite{finn2017model} introduced a meta-learning method based on learning easily adaptable model parameters through gradient descent. However, these methods aim to automatically learn the better first-order similarity by optimizing hyper-parameters while the proposed method is to learn second-order similarity.

\section{Second-Order Similarity}
\begin{figure}
  \centering
  \includegraphics[width=1.0\textwidth]{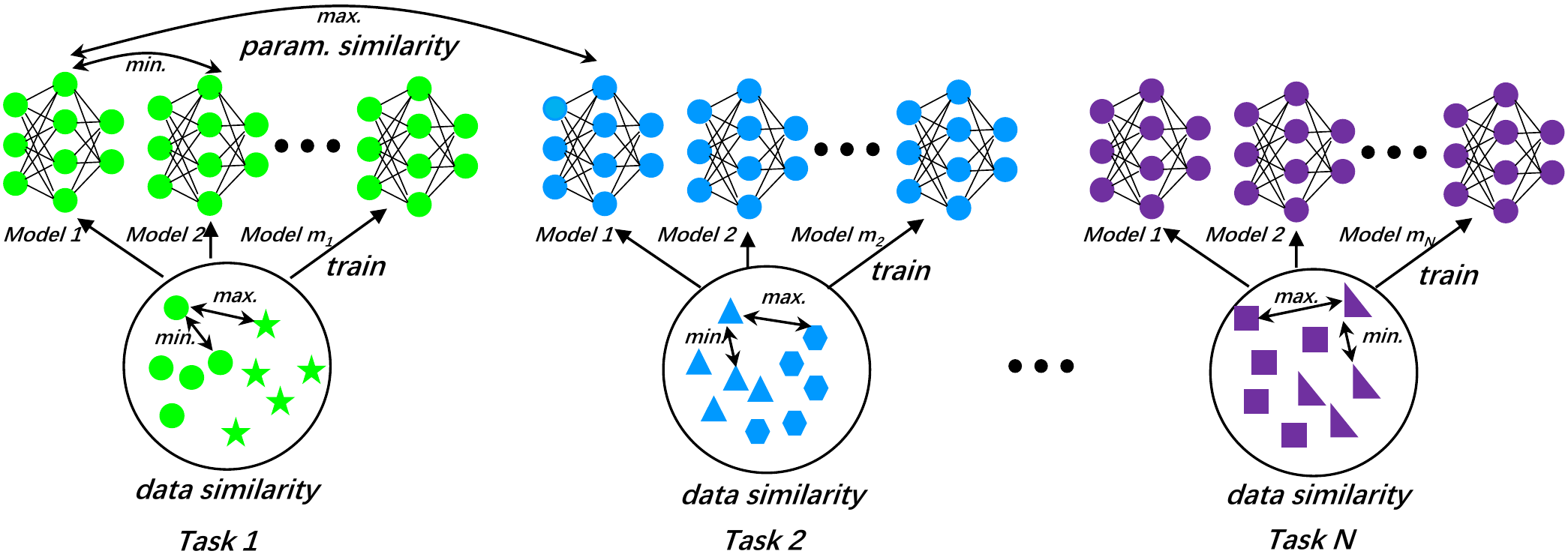}\\
  \caption{Data similarity and parameter similarity}\label{fig:problem}
\end{figure}

We first consider data similarity, which is also termed first-order similarity in this paper. For a learning task, let $D$ denote a training dataset with $n$ data points $\{x_i, y_i\}_{i=1}^{n}$. Let $f(D;{\theta})$ be an objective function $f$ with parameters $\theta$ on $D$. The objective function is optimized by the standard gradient descent as follows
\begin{equation}
\label{eq:first_order}
{\theta _{t + 1}} = {\theta _t} - {\alpha _t}\nabla f(D;{\theta _t})
\end{equation}
where $\alpha _t$ is a step size multiplier at $t$. After a sequence of updates, the algorithm converges to a minimizer $\theta^{*}=argmin_{\theta}f(D;\theta)$. With the parameter $\theta^{*}$, the learning task can measure similarity between data points. In this way, given a set of learning tasks, we can obtain a set of trained task-specific parameters. Now the question is how to measure the similarity between different task-specific parameters?

Data similarity/metric learning aims to minimize the distance between intra-class data points and maximize the distance between inter-class data points. Similar to data similarity learning, the goal of parameter similarity/metric learning is to minimize the distance between intra-task models and maximize the distance between inter-task models, as illustrated in Figure \ref{fig:problem}. Intra-task models are obtained by repeating the training procedure many times for each task.

The intra-task models are diverse and contain rich statistical semantics due to the intractable non-convexity of deep networks and the coupled relationship between data points and parameters in convolution operation. Data points and parameters share the same semantics. For example, car classification and bus classification share a similar model while car classification and bird classification could produce dissimilar models. That is, the distance between car classification and bus classification models is closer than that between car classification and bird classification models. Learning to distinguish intra-task and inter-task parameters can uncover the underlying parameter patterns of local solutions and thus provides a parameter similarity metric for task relation.

Let $\mathcal{T}$ denote a set of tasks $\{\mathcal{T}_i\}_{i=1}^{N}$. For $\mathcal{T}_i\in \mathcal{T}$, we train $m_i$ deep models with a task label $\mathcal{Y}_i$. We obtain a model set $\mathcal{D}$ with $M$ models $\{(\theta^{*}_j, \mathcal{Y}_j)\}_{j=1}^{M}$ from $N$ tasks, where $M = \sum\limits {{m_j}}$. $\theta^{*}_j$ is a trained model, which can be regarded as a metadata point. We then use these $M$ metadata points to train the second-order similarity learning by
\begin{equation}
\label{eq:second_order}
{\phi _{t + 1}} = {\phi _t} - \beta_t \nabla h(\mathcal{D};{\phi _t})
\end{equation}
where $h$ is a second-order similarity objective function with parameters $\phi$ on $\mathcal{D}$ and $\beta_t$ is a step size multiplier at $t$.

Second-order similarity learning is naturally different from existing learning to learn methods. First, the goal of second-order similarity learning is to learn parameter similarity based on first-order similarity while learning to learn methods learn hyper-parameters of an optimizer to navigate data similarity learning, which is naturally the first-order similarity learning. Second, second-order similarity learning optimizes model parameters and second-order model parameters (different from ``hyper-parameter") separately while learning to learn methods optimize both model parameters and hyper-parameters simultaneously. Such a disjoint optimization method guarantees that trained deep models have converged to a good local solution.

\section{Second-Order Neural Networks}
\begin{figure}
  \centering
  \includegraphics[width=1.0\textwidth]{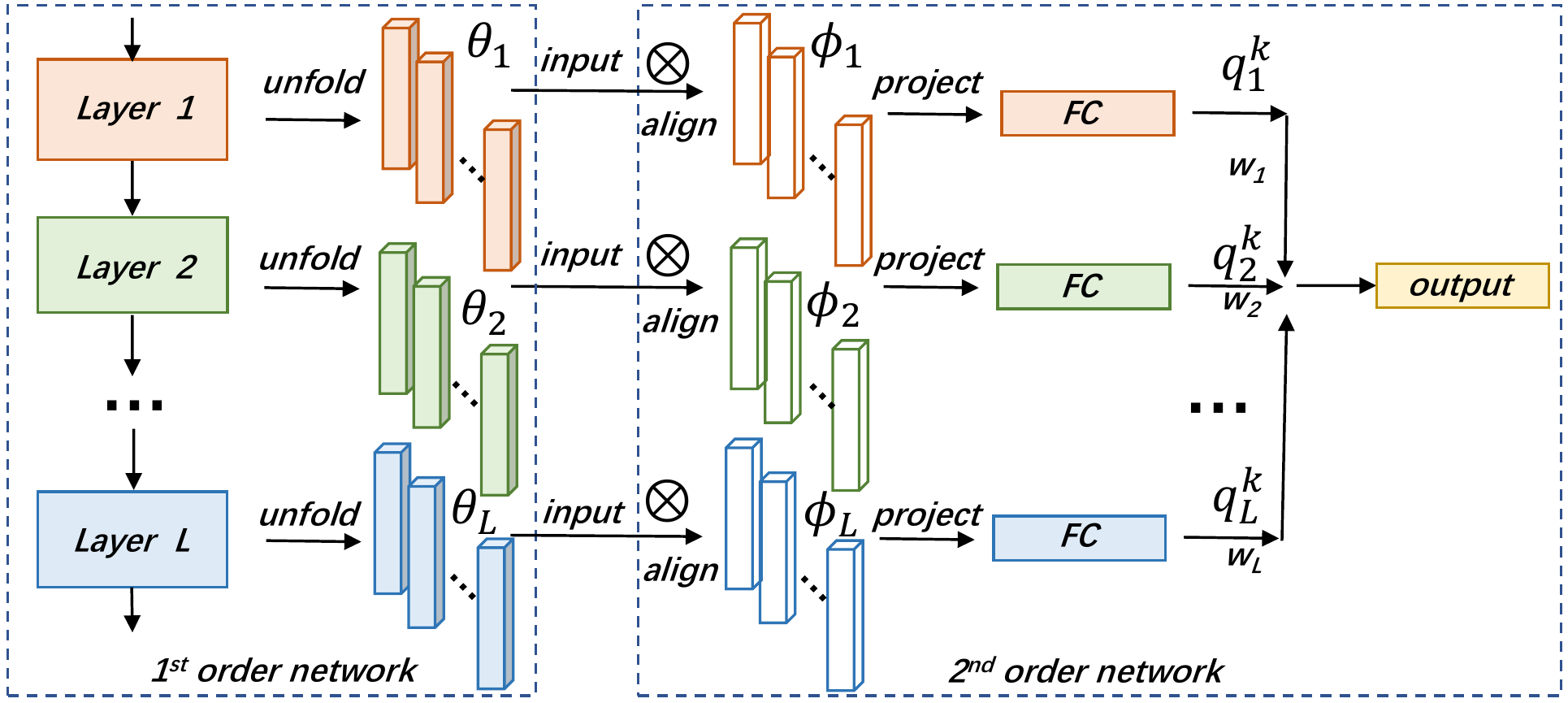}\\
  \caption{Second-order neural network}\label{fig:overview}
\end{figure}
In this section, we introduce a hierarchical second-order neural network to learn an effective parameter representation for second-order similarity, as shown in Figure \ref{fig:overview}. Specifically, the second-order network consists of $L$ branches to deal with $L$ semantic levels of $L$ layers. For each branch, we attempt to align metadata and then project it into a common space for parameter similarity learning with a fully connected layer. Finally, we weight the losses of $L$ different branches to control the relative importance.

Although the idea of parameter similarity learning is similar to data similarity learning, it is a challenging problem because of high-dimensional metadata points (higher than images, e.g., 96M for ResNet-50), changeable order of filters and their channels at a layer, and hierarchical semantics.

One of the most challenging problems is that the order of filters and their channels is changeable when repeating the training procedure. Therefore, directly using fully connected layers to project these metadata points into a common space cannot obtain good performance. We prove that the order of filters and their channels at a layer could be shuffled during different training procedures as follows.

Consider the convolution operation in visual tasks. Let $F_l$ be feature maps of size $(c_l, H_l, W_l)$ at the $l$-th convolutional layer. $F_l^c$ is the $c$-th feature map of $F_l$ ($c \in [1,c_l]$). $\odot$ is the convolution operator. We re-use $\theta _l$ as filters of size $(g_l,c_l,h_l,w_l)$ at the $l$-th convolutional layer for simplification (do not be confused by $\theta _l$, $\theta_j^{*}$ and $\theta_t$ mentioned above). $\theta _l^g$ is the $g$-th filter of $\theta _l$ ($g\in [1,g_l]$). $\theta_l^{g,c}$ is the $c$-th channel of $\theta _l^g$. $\forall g \in [1,g_l]$, $\theta _l^g$ produces a feature map $F_{l + 1}^g$. The convolution operation is given by
\begin{equation}
\label{eq:conv1}
{{F_l} \odot \theta _l^g = F_{l + 1}^g}
\end{equation}
In this way, $g_l$ filters produce $g_l$ feature maps as output. We can see that the convolution operation is independent of the order of $g_l$ filters $\{\theta _l^g\}_{g=1}^{g_l}$. Although $g_l$ shuffled filters at the $l$-layer naturally lead to $g_l$ shuffled channels of $F_l$, it does not reduce informative features. The convolution operation for each filter at a layer is symmetric. Therefore, the order of $\theta _l$ is changeable. We perform the next convolution operation by
\begin{equation}
\label{eq:conv2}
{{F_{l + 1}} \odot \theta _{l+1}^g = F_{l + 2}^g}
\end{equation}
According to Eq. (\ref{eq:conv1}), shuffled ${\{\theta_l^g\}}_{g=1}^{g_l}$ leads to shuffled ${\{F_{l+1}^g\}}_{g=1}^{g_l}$, where ${g_l}=c_{l+1}$. According to Eq. (\ref{eq:conv2}), the order of $g_{l+1}$ channels of $F_{l + 2}^g$ is determined by the order of ${\{F_{l+1}^g\}}_{g=1}^{g_l}$ and the channel order of the filter $\theta _{l+1}^g$. Because the order of ${\{F_{l+1}^g\}}_{g=1}^{g_l}$ is changed, the channel order of the filter $\theta _{l+1}^g$ have to be changed if we fix the channel order of the output $F_{l + 2}$. For example, if we exchange the order of two filters $\theta_l^{g}$ and $\theta_l^{g^{'}}$, and also exchange the channel $g$ and $g^{'}$ of $\theta_{l+1}^{j}$ for any $j\in [1,g_l]$, the output of the network is fixed. That is, there are a large number of filter map combinations such that the output is the same.

Let $\mathcal{O}_l^g$ and $\mathcal{O}_l^c$ denote the filter order and channel order of $\theta_l$. We denote the order of $\theta_l$ as $(\mathcal{O}_l^c,\mathcal{O}_l^g)$. We define the \textbf{order chain rule} of convolutional filters as follows
\begin{itemize}
  \item If $l=1$, then $\mathcal{O}_l^c$ is fixed because it is constrained by the channels of natural images, e.g., RGB. $\mathcal{O}_l^g$ determines $\mathcal{O}_{l+1}^c$
  \item If $l\geq2$, then $\mathcal{O}_l^c$ is determined by $\mathcal{O}_{l-1}^g$ while $\mathcal{O}_l^g$ determines $\mathcal{O}_{l+1}^c$.
  \item If $l=L$, then $\mathcal{O}_l^c$ is determined $\mathcal{O}_{l-1}^g$. $\mathcal{O}_l^g$ is fixed because it is constrained by the loss of neural networks (or labels).
\end{itemize}

The order chain rule provides an ideal case that intra-task models are only affected by the changeable order of filters and their channels. In fact, intra-task deep models are also affected by hierarchical semantics and non-convex optimization. Therefore, give two deep models with a shared task label, it is difficult to align the changeable order of filters and their channels, which confuses parameter similarity learning.

To address this problem, we propose to transform $(\mathcal{O}_l^c,\mathcal{O}_l^g)$ of different intra-task models into a ``standard" order to align these model parameters. Let ${\phi}_l$ is the second-order filters of size $(g_l^{'},c_l,h_l,w_l)$. ${\theta}_l$ and ${\phi}_l$ could have different number filters at a layer. $\forall {c_i},{c_j}\in [1,c_l]$, $\forall g_i\in [1,g_l]$, $\forall g_j^{'} \in [1,{g_l^{'}}]$ and $\forall l\in[2,L]$, we first compute the Frobenius inner product between filter maps and second-order filter maps by
\begin{equation}
\label{eq:map2map}
s_l^{g_i,g_j,{c_i},{c_j}} = \theta _l^{g_i,{c_i}} \otimes \phi _l^{g_j^{'},{c_j}}
\end{equation}
where $\otimes$ denotes the Frobenius inner product. In Eq. (\ref{eq:map2map}), we compute the Frobenius inner product between all pairs of $\theta _l^{g_i}$ and $\phi _l^{g_j^{'}}$ filter maps. Without any prior information, we cannot directly compute the standard $\mathcal{O}_l^c$. We estimate the standard $\mathcal{O}_l^c$ based on the max matching between $\theta _l^{g_i}$ and $\phi _l^{g_j^{'}}$. We have
\begin{equation}
\label{eq:max}
s_l^{g_i,g_j^{'}} = \sum\limits_{{c_j}} {\mathop {\max }\limits_{{c_i}} s_l^{g_i,g_j^{'},{c_i},{c_j}}}
\end{equation}
When $l=1$, we do not need to align $\mathcal{O}_l^c$ by the max operation. $\forall {c}\in [1,c_l]$, $\forall g_i\in [1,g_l]$, and $\forall g_j^{'} \in [1,{g_l^{'}}]$, we directly compute $s_l^{g_i,g_j^{'}}$ by
\begin{equation}
\label{eq:map2map2}
s_l^{g_i,g_j^{'}} = \theta _l^{g_i,{c}} \otimes \phi _l^{g_j^{'},{c}}
\end{equation}
$\forall l\in [1,L-1]$, we then estimate the standard $\mathcal{O}_l^g$ by using the sort operation
\begin{equation}
\label{eq:sort}
s_l^{{r_1},{g_j}},s_l^{{r_2},{g_j}},..,s_l^{{r_{{g_l}}},{g_j}} = sort(s_l^{1,{g_j}},s_l^{2,{g_j}},..,s_l^{{g_l},{g_j}})
\end{equation}
where $r_1,r_2,...,r_{g_l}$ is the new filter order indexes. Based on the order chain rule, $\mathcal{O}_l^c$ is fixed when $l=L$. Therefore, we do not need to align $\mathcal{O}_l^g$.

After aligning $\mathcal{O}_l^g$, we add a fully connected layer to project the aligned parameter representation into common space. Finally, we train a second-order model by using a conventional metric learning loss, e.g., cross-entropy loss \cite{sun2014deep} or triplet loss \cite{ding2015deep}. In this paper, we simply use the cross-entropy loss for model similarity measure
\begin{equation}
    \label{eq:loss}
    {\mathcal{L}} =  - \sum\limits_{i = 1}^N {{{\mathcal{Y}_i}}\log( {\frac{1}{W}\sum\limits_{l = 1}^L{w_lq_i^l}})}
\end{equation}
where ${{\mathcal{Y}_i}}$ is the $i$-dimensional value of the one-hot label $\mathcal{Y}$. $q_i^l$ represents the probability of the $i$-th visual task at the $l$-th branch. $w_l$ controls the relative importance of $L$ branches. $W = \sum\limits_{l = 1}^L {{w_l}}$. When measuring the parameter similarity between two deep models, we extract parameter features from the last fully connected layer of the second-order network. We normalize parameter features and use the cosine similarity to compute the parameter similarity. 

\section{Experiments}
\label{sec:exp}
In this section, we conduct extensive experiments to validate the effectiveness of our proposed method. All experiments are conducted with Pytorch \cite{paszke2017automatic}.

\textbf{\emph{Dataset and Modelset.}} The CIFAR-100 dataset \cite{krizhevsky2009learning} has 100 classes containing 600 images each. There are 500 training images and 100 testing images per class.

To measure model-based parameter similarity, we create a model set called ModelSet500 based on CIFAR-100. Specially, we split 100 classes into 50 groups as 50 visual tasks with task labels 0$\sim$49. Each task contains 2 classes. For each task, we repeat the training procedure 10 times to obtain 10 deep models. Finally, we obtain 500 deep models for 50 tasks.

\textbf{\emph{First-order and Second-order Network Implementation.}} We implement a ResNet-20 network \cite{he2016deep} as the first-order network in our experiments. The network consists of one convolutional layer, three residual blocks, one global average pooling layer, and one fully-connected layer. The blocks of ResNet-20 consist of 6 convolutional layers, respectively (excluding $1\times 1$ convolutional layers which are used for downsampling in residual blocks). When computing the parameter similarity measure, we omit the last fully-connected layer and $1\times 1$ convolutional layers. We use SGD with a mini-batch size of 128. The learning rate starts from 0.1 and is divided by 10 after 80 epochs and 120 epochs, respectively. We train ResNet-20 for 160 epochs. We use a weight decay of 0.0001 and a momentum of 0.9. We set the size of padding to 4 and perform random cropping with $32\times 32$ and random horizontal flip. 

When implementing the second-order network, we use SGD with a mini-batch size 1. The learning rate starts from 0.001 and is divided by 10 after 40 epochs and 80 epochs, respectively. We train 100 epochs. We set the loss weights $w_1=1.0$ and $w_2=w_3=...=w_9=0.0$ in Section \ref{subsec:task_classification}\, \ref{subsec:retrieval} and \ref{subsc:transfer} because the first branch plays a major role in Eq. (\ref{eq:loss}). We analyze the performance of different branches in Section \ref{subsc:layerwise} and \ref{subsc:alignment}.


\subsection{Task classification}
\label{subsec:task_classification}
We first adopt the conventional image classification evaluation metric for task classification on ModelSet500. The goal of task classification is to predict task labels of trained models. For each task, we sample eight deep models for training while the other two for test. Therefore, the training set contains 400 deep models while the test set contains 100 deep models. We train the second-order network by classifying 50 task classes.

To evaluate the effectiveness of the second-order network, we set three baselines, i.e., random prediction without using any second-order model, only fully connected layer, Frobenius inner product without aligning filters and their channels. Compared with these three baselines, we can see that the proposed method achieves better performance. As shown in Table \ref{tab:classification}, our method achieves 79.8\% top-1 accuracy while the three baselines are lower than 6\%. The main reason may be that the changeable order of filters and their channels confuses the second-order neural networks.

\begin{table*}[]
\caption{Task classification.}\label{tab:classification}
\begin{center}
\begin{tabular}{c|c|c|c}
\hline
methods     & top-1 (\%) & top-5 (\%) & top-10 (\%) \\ \hline
random prediction      & 2.0   & 9.6   & 18.3   \\ \hline
only one FC layer      & 5.3   & 17.5   & 29.3   \\ \hline
Frobenius+FC,w/o alignment & 5.5   & 18.3    & 29.0     \\ \hline
\textbf{Frobenius+FC, w/ alignment (Ours)} & \textbf{79.8}  & \textbf{94.8}  & \textbf{98.0}   \\ \hline
\end{tabular}
\end{center}
\end{table*}

\subsection{Task retrieval}
\label{subsec:retrieval}
We then adopt the conventional image retrieval evaluation metric for task retrieval on ModelSet500. Given a query deep model, the goal of task retrieval is to retrieve the target deep models from a gallery. We split ModelSet500 into a training set and a test set. The training set includes 40 tasks and the test set includes 10 tasks. For each task in the test set, we sample two deep models for query models while the others are used for gallery models. Task retrieval is challenging and practical in real-world scenarios. It can be used to retrieve similar visual tasks for transfer learning. Different from task classification, we train the second-order network by classifying 40 task classes and extract the fully connected layer as parameter representation during the test phase. We use the normalized parameter representation for task retrieval.

We also set three baselines like task classification. Compared with these three baselines, we can see that the proposed method achieves better performance. As shown in Table \ref{tab:retrieval}, our method achieves 80.7\% rank-1 accuracy while the three baselines are lower than 15\%. The reason is the same as Section \ref{subsec:task_classification}.

\begin{table*}[]
\caption{Task retrieval.}\label{tab:retrieval}
\begin{center}
\begin{tabular}{c|c|c|c}
\hline
methods     & rank-1 (\%) & rank-5 (\%) & rank-10 (\%) \\ \hline
random prediction     & 10.0   & 41.0   & 65.1   \\ \hline
only one FC layer      & 15.0   & 55.8   & 83.3    \\ \hline
Frobenius+FC, w/o alignment & 13.3   & 58.3    & 80.0     \\ \hline
\textbf{Frobenius+FC, w/ alignment (Ours)} & \textbf{80.7}  & \textbf{95.0}  & \textbf{95.8}    \\ \hline
\end{tabular}
\end{center}
\end{table*}
\begin{figure}[!htp]
\centering
\subfloat[Task transferability.]{\includegraphics[width=0.3\textwidth]{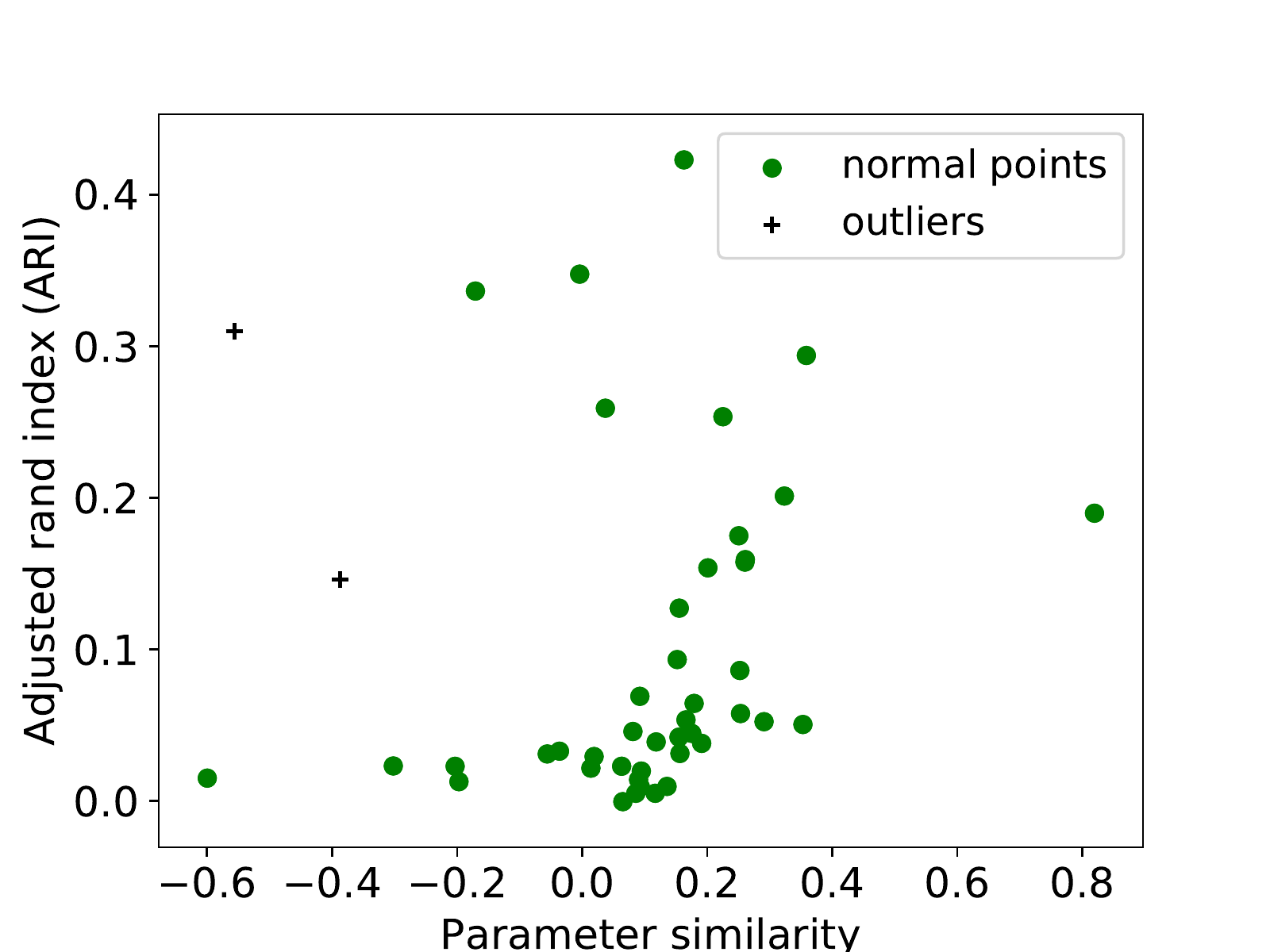}
\label{fig:transfer}}
\hfil
\subfloat[Effect of different branches.]{\includegraphics[width=0.3\textwidth]{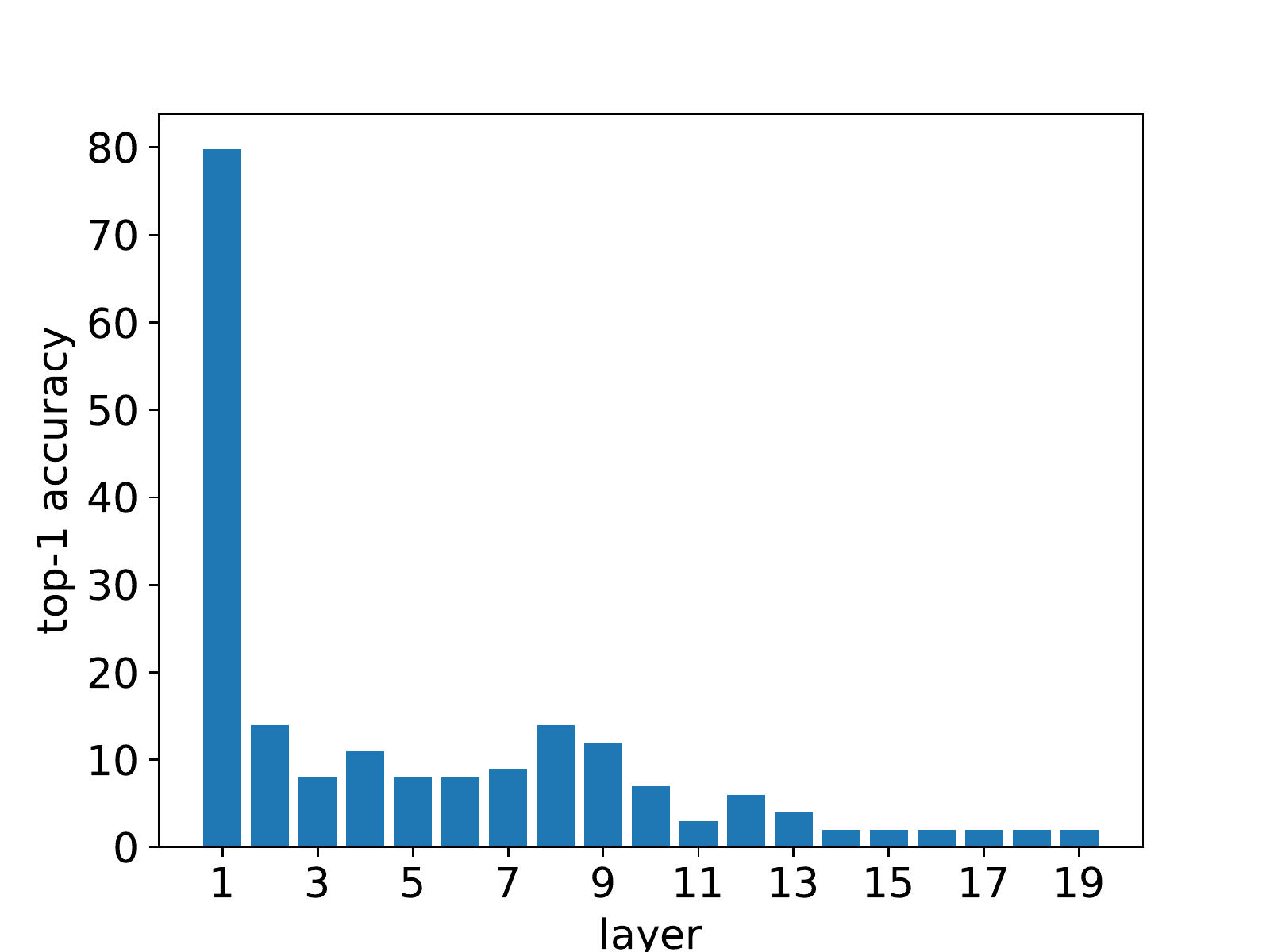}  
\label{fig:layerwise}}
\hfil
\subfloat[Effectiveness of the alignment methods.]{\includegraphics[width=0.3\textwidth]{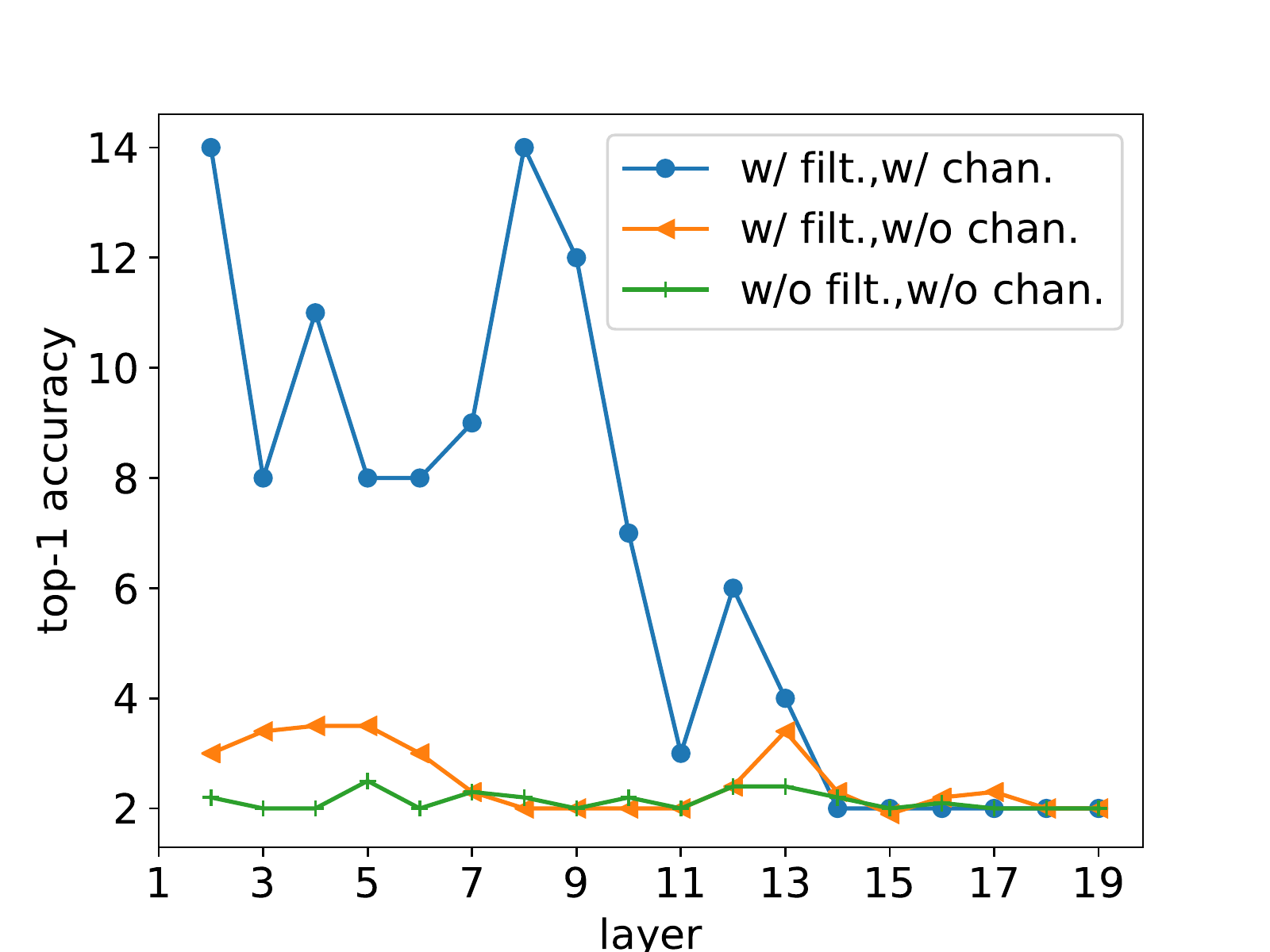}
\label{fig:comp}}
\hfil
\caption{The effectiveness of learnable parameter similarity.}
\label{fig:continuity}
\end{figure}
\subsection{Task transferability}
\label{subsc:transfer}
To validate that the parameter similarity can be used to measure task transferability, we conduct an experiment to study task transferability with respect to parameter similarity. We use the task retrieval setting to obtain parameter representation, as discussed in Section \ref{subsec:retrieval}. We evaluate task transferability by training first-order models for one task (a labeled source domain) and applying these models to another task (an unlabeled target domain) for data feature extraction and k-means clustering. We use adjusted rand index (ARI) \cite{vinh2010information} for evaluation, which is widely used in cluster analysis.

As shown in Figure \ref{fig:transfer}, although there are some outliers, we observe that parameter similarity is roughly proportional to task transferability (ARI). In most of the cases, the higher parameter similarity score the source and target tasks have, the better task transferability they can achieve.

\subsection{Effect of different branches}
\label{subsc:layerwise}
To show the relative importance of different branches for the parameter similarity measure, we conduct an experiment on ModelSet500 by keeping one of $L$ branches while removing the other $L-1$ branches. That is, when evaluating the $l^{'}$-th branch, we set $w_l=1.0$ if $l=l^{'}$, otherwise $w_l=0.0$. We use the task classification setting. Different from other experiments, the learning rate in this experiment is divided by 10 after 150 epochs and 225 epochs, respectively. We train each branch for 300 epochs because it takes more time to align the order of both filters and their channels.

As shown in Figure \ref{fig:layerwise}, it is observed that the first branch is the most important branch and it can achieve 79.8\% top-1 accuracy. From the 2rd to 13th branch, the second-order network still can learn a weak discriminative parameter representation since the results are higher than top-1 accuracy of random prediction, i.e., 2.0\% (50 classes). However, from the 13th to 19th layer, the second-order cannot learn a discriminative parameter representation since the results are nearly reduced to 2.0\% top-1 accuracy. The reason could be that the order chain rule leads to the cumulative error of $(\mathcal{O}_l^c,\mathcal{O}_l^g)$ alignment at higher layers.

Besides, it is still hard to fuse all of the branches to improve better performance because the 2rd to 19th branch is much lower. This is the reason why we only use the first branch for task classification and retrieval.

\subsection{Effectiveness of the alignment methods}
\label{subsc:alignment}
To show the effectiveness of the alignment methods, we conduct an experiment by isolating different alignment methods, i.e., filter alignment (filt.) and channel alignment (chan.). The training setting is the same as Section \ref{subsc:layerwise}. We only analyze the 2rd to 19th branch because the 1st branch only contains one unaligned case, i.e., unaligned filters (see Eq. (\ref{eq:map2map2})). As shown in Figure \ref{fig:comp}, it is observed that both alignment methods are important for parameter similarity learning.
\section{Advantages and disadvantages}
The proposed method comes with advantages and disadvantages compared with the existing methods. The advantages are that the learnable parameter similarity model does not need to jointly train a large number of transfer networks to estimate task relations. The second-order network does not need to align convolutional filters if the first-order network does not contain convolutional layers in some non-image scenarios. Learnable parameter similarity can be also used to quickly retrieve related tasks from the model market (set) for transfer learning.

The disadvantages are that the proposed learnable parameter similarity cannot measure the similarity between different networks and it is still hard to fuse all of the branches to improve better performance. The reason could be that the alignment at higher layers are still unsatisfactory.
\section{Conclusion}
In this paper, we present a Learnable Parameter Similarity (LPS) method to reveal the underlying relations between different visual tasks. Our approach learns task relations by using task-specific trained deep models instead of jointly training a large number of transfer networks. We present a hierarchical second-order network to deal with high-dimensional unaligned deep models. We evaluate the LPS method on a new parameter similarity learning benchmark ModelSet500 and extensive experiments show the effectiveness of the proposed method. Future work is to further explore the fusion of different layers and the alignment of the convolutional filters at higher layers and develop a parameter similarity metric for heterogeneous networks.

\medskip

\small

\bibliography{neurips2019}

\begin{thebibliography}{10}

\bibitem{andrychowicz2016learning}
Marcin Andrychowicz, Misha Denil, Sergio Gomez, Matthew~W Hoffman, David Pfau,
  Tom Schaul, Brendan Shillingford, and Nando De~Freitas.
\newblock Learning to learn by gradient descent by gradient descent.
\newblock In {\em Advances in Neural Information Processing Systems}, pages
  3981--3989, 2016.

\bibitem{bengio2011deep}
Yoshua Bengio.
\newblock Deep learning of representations for unsupervised and transfer
  learning.
\newblock In {\em Proceedings of the 2011 International Conference on
  Unsupervised and Transfer Learning workshop-Volume 27}, pages 17--37. JMLR.
  org, 2011.

\bibitem{ding2015deep}
Shengyong Ding, Liang Lin, Guangrun Wang, and Hongyang Chao.
\newblock Deep feature learning with relative distance comparison for person
  re-identification.
\newblock {\em Pattern Recognition}, 48(10):2993--3003, 2015.

\bibitem{finn2017model}
Chelsea Finn, Pieter Abbeel, and Sergey Levine.
\newblock Model-agnostic meta-learning for fast adaptation of deep networks.
\newblock In {\em Proceedings of the 34th International Conference on Machine
  Learning-Volume 70}, pages 1126--1135. JMLR. org, 2017.

\bibitem{he2016deep}
Kaiming He, Xiangyu Zhang, Shaoqing Ren, and Jian Sun.
\newblock Deep residual learning for image recognition.
\newblock In {\em Proceedings of the IEEE conference on computer vision and
  pattern recognition}, pages 770--778, 2016.

\bibitem{hoffman2017cycada}
Judy Hoffman, Eric Tzeng, Taesung Park, Jun-Yan Zhu, Phillip Isola, Kate
  Saenko, Alexei~A Efros, and Trevor Darrell.
\newblock Cycada: Cycle-consistent adversarial domain adaptation.
\newblock {\em arXiv preprint arXiv:1711.03213}, 2017.

\bibitem{krizhevsky2009learning}
Alex Krizhevsky and Geoffrey Hinton.
\newblock Learning multiple layers of features from tiny images.
\newblock Technical report, Citeseer, 2009.

\bibitem{li2009transfer}
Bin Li, Qiang Yang, and Xiangyang Xue.
\newblock Transfer learning for collaborative filtering via a rating-matrix
  generative model.
\newblock In {\em Proceedings of the 26th annual international conference on
  machine learning}, pages 617--624. ACM, 2009.

\bibitem{liang2018m2m}
Wenqi Liang, Guangcong Wang, Jianhuang Lai, and Junyong Zhu.
\newblock M2m-gan: Many-to-many generative adversarial transfer learning for
  person re-identification.
\newblock {\em arXiv preprint arXiv:1811.03768}, 2018.

\bibitem{liu2018progressive}
Chenxi Liu, Barret Zoph, Maxim Neumann, Jonathon Shlens, Wei Hua, Li-Jia Li,
  Li~Fei-Fei, Alan Yuille, Jonathan Huang, and Kevin Murphy.
\newblock Progressive neural architecture search.
\newblock In {\em Proceedings of the European Conference on Computer Vision
  (ECCV)}, pages 19--34, 2018.

\bibitem{liu2018darts}
Hanxiao Liu, Karen Simonyan, and Yiming Yang.
\newblock Darts: Differentiable architecture search.
\newblock {\em arXiv preprint arXiv:1806.09055}, 2018.

\bibitem{liu2016coupled}
Ming-Yu Liu and Oncel Tuzel.
\newblock Coupled generative adversarial networks.
\newblock In {\em Advances in neural information processing systems}, pages
  469--477, 2016.

\bibitem{long2017deep}
Mingsheng Long, Han Zhu, Jianmin Wang, and Michael~I Jordan.
\newblock Deep transfer learning with joint adaptation networks.
\newblock In {\em Proceedings of the 34th International Conference on Machine
  Learning-Volume 70}, pages 2208--2217. JMLR. org, 2017.

\bibitem{pan2010survey}
Sinno~Jialin Pan and Qiang Yang.
\newblock A survey on transfer learning.
\newblock {\em IEEE Transactions on knowledge and data engineering},
  22(10):1345--1359, 2010.

\bibitem{paszke2017automatic}
Adam Paszke, Sam Gross, Soumith Chintala, Gregory Chanan, Edward Yang, Zachary
  DeVito, Zeming Lin, Alban Desmaison, Luca Antiga, and Adam Lerer.
\newblock Automatic differentiation in pytorch.
\newblock 2017.

\bibitem{rosenstein2005transfer}
Michael~T Rosenstein, Zvika Marx, Leslie~Pack Kaelbling, and Thomas~G
  Dietterich.
\newblock To transfer or not to transfer.
\newblock In {\em NIPS 2005 workshop on transfer learning}, volume 898, pages
  1--4, 2005.

\bibitem{sun2014deep}
Yi~Sun, Xiaogang Wang, and Xiaoou Tang.
\newblock Deep learning face representation from predicting 10,000 classes.
\newblock In {\em Proceedings of the IEEE conference on computer vision and
  pattern recognition}, pages 1891--1898, 2014.

\bibitem{tzeng2017adversarial}
Eric Tzeng, Judy Hoffman, Kate Saenko, and Trevor Darrell.
\newblock Adversarial discriminative domain adaptation.
\newblock In {\em Proceedings of the IEEE Conference on Computer Vision and
  Pattern Recognition}, pages 7167--7176, 2017.

\bibitem{vinh2010information}
Nguyen~Xuan Vinh, Julien Epps, and James Bailey.
\newblock Information theoretic measures for clusterings comparison: Variants,
  properties, normalization and correction for chance.
\newblock {\em Journal of Machine Learning Research}, 11(Oct):2837--2854, 2010.

\bibitem{zamir2018taskonomy}
Amir~R Zamir, Alexander Sax, William Shen, Leonidas~J Guibas, Jitendra Malik,
  and Silvio Savarese.
\newblock Taskonomy: Disentangling task transfer learning.
\newblock In {\em Proceedings of the IEEE Conference on Computer Vision and
  Pattern Recognition}, pages 3712--3722, 2018.

\bibitem{zoph2016neural}
Barret Zoph and Quoc~V Le.
\newblock Neural architecture search with reinforcement learning.
\newblock {\em arXiv preprint arXiv:1611.01578}, 2016.

\end{thebibliography}
\bibliographystyle{plain}

\end{document}